\documentclass[11pt]{article}
\usepackage[preprint]{acl}
\usepackage{times}
\usepackage{latexsym}
\usepackage[T1]{fontenc}
\usepackage[utf8]{inputenc}
\usepackage{microtype}
\usepackage{comment}
\IfFileExists{inconsolata.sty}{\usepackage{inconsolata}}{}
\usepackage{booktabs,multirow,array}

\usepackage{graphicx}
\usepackage{amsmath,amssymb}
\usepackage{xcolor}
\usepackage{tikz}
\usetikzlibrary{arrows.meta,positioning,fit,backgrounds}
\usetikzlibrary{positioning,arrows.meta}
\usepackage{listings}
\usepackage{newfloat}
\DeclareFloatingEnvironment[fileext=lol,listname={List of Listings},name=Listing,placement=tbp]{listing}
\newcommand{\src}[1]{}

\title{Definitions, Label Names, or Demonstrations?\\
Auditing What Governs LLM Annotation}

\author{Viraj Mishra \quad Shubham Gupta \quad Dhruv Gupta \\
    Department of Computer Science and Engineering \\
    Pennsylvania State University
    \AND
    Yuehong Cassandra Tai\thanks{\hspace{1pt}Corresponding authors.} \\
    Center for Social Data Analytics \\
    Pennsylvania State University \\
    \texttt{yhcasstai@psu.edu}
    \And
    Qunhua Li\footnotemark[1] \\
    Department of Statistics \\
    Pennsylvania State University \\
    \texttt{qul12@psu.edu}}

\begin{document}
\maketitle

\begin{abstract}
Large language models (LLMs) are increasingly used to annotate social-science texts, but high accuracy does not establish that they apply researcher-supplied category definitions: predictions may instead reflect prior associations with category labels or labeled demonstrations. We introduce a counterfactual audit that reassigns definitions among output labels, distinguishing adherence to supplied definitions from preservation of a dataset's original mapping. Across three political-text datasets, models rarely follow reassigned definitions, and standard fine-tuning strengthens the original mapping without reliably increasing definition adherence. We then study retrieval-augmented annotation. Retrieving the nearest training examples generally outperforms retrieving examples separately within each category despite using fewer demonstrations. On the two tasks with substantial gains, direct \(k\)-nearest-neighbor classification recovers 74--93\% of the improvement, indicating that neighborhood labels provide much of the predictive signal. Relabeling fixed demonstrations redirects predictions, whereas aligning demonstrations with reassigned definitions substantially increases adherence to the new mapping. Thus, accuracy-improving interventions need not increase codebook adherence; definitions, category labels, demonstrations, retrieval, and training data should be aligned and validated jointly. Code and data for all experiments are available at \url{https://anonymous.4open.science/r/llm-annotation-FD2D}.
\end{abstract}

\section{Introduction}

Social scientists use annotation codebooks to translate theoretical concepts into observable categories in text. The resulting annotations measure substantively important constructs, such as forms of protest, types of political violence, and policy positions in party platforms \citep{grimmer2022text}. Large language models (LLMs) offer a scalable approach to producing such annotations and can approach or exceed crowd-worker performance on many classification tasks \citep{gilardi2023chatgpt,ziems2024can,tornberg2025large}. Yet predictive accuracy alone leaves a fundamental question unanswered: \emph{what annotation rule is the model actually applying?}

A correct prediction may reflect the supplied definition in the codebook, prior associations with a category label, or patterns in labeled demonstrations. Because these sources normally imply the same answer, conventional evaluation cannot distinguish them. This matters for construct validity: a model may reproduce historical labels while failing to implement the researcher’s intended operationalization, particularly when definitions change or prior annotations depart from the written rules. 

Prior work shows that category labels, definitions, and demonstrations all affect LLM classification. Replacing informative labels with generic identifiers, swapping label--definition pairings, or ablating codebook components can substantially change performance \citep{halterman2026codebook}, and models use definitions and demonstrations unevenly \citep{milios2023,mohammadi2025llms}. We ask whether task-specific fine-tuning increases responsiveness to supplied definitions or strengthens reproduction of the original mapping, and how predictions respond when definition and demonstration mappings are varied independently.

We study how LLM annotators combine three sources of guidance: (1) category labels and their prior associations, (2) written definitions, and (3) labeled demonstrations. Our counterfactual audit reassigns definitions among output labels, placing the supplied definition mapping in conflict with the dataset's original mapping. We use it to test whether predictions follow the definitions, preserve the original mapping, or follow neither, and whether fine-tuning changes this behavior. We extend the framework to retrieval-augmented annotation. We compare \emph{Global Example-RAG}, which retrieves the nearest examples across the training pool, with \emph{Class-wise Example-RAG}, which retrieves examples separately within each category. Holding the retrieved texts fixed, we then independently manipulate definition and demonstration-label mappings to test which mapping governs predictions.

We ask four questions. \textbf{Q1:} When supplied definitions conflict with the original mapping, which mapping do LLM annotators follow? \textbf{Q2:} Do accuracy-improving interventions, including fine-tuning and retrieval, increase adherence to supplied definitions? \textbf{Q3:} How do retrieval strategy and neighborhood label composition affect annotation accuracy? \textbf{Q4:} When definitions and retrieved demonstrations encode conflicting mappings, which source governs predictions?

Across three political-text codebooks, models rarely treat reassigned definitions as authoritative, and fine-tuning reinforces the original mapping without reliably increasing definition adherence. On the more complex tasks, Global Example-RAG outperforms both Class-wise Example-RAG and annotation without retrieval, but direct nearest-neighbor classification recovers much of the gain. Relabeling fixed demonstrations redirects predictions; written definitions moderate but do not eliminate this influence. Thus, higher accuracy need not imply more faithful application of the supplied codebook.

We contribute a behavioral audit that separates definition adherence from preservation of the original mapping; controlled analyses of retrieval architecture, neighborhood composition, and demonstration mappings; and practical guidance for aligning and validating the label system, codebook, retrieval procedure, and demonstration bank as a joint measurement instrument.

\section{Related Work}
\label{sec:related}

\paragraph{LLM annotation in social science.}
Studies of LLM annotation primarily evaluate agreement with human judgments across tasks and prompts \citep{gilardi2023chatgpt,ziems2024can,pangakis2023automated,abraham2025prompt}; practical guidance emphasizes clear tasks and validation against human-coded texts \citep{tornberg2024best,alizadeh2025}. These comparisons establish predictive performance, but agreement with human labels does not reveal whether a model follows the supplied coding rules. We examine this question by placing category-label associations, definitions, and demonstrations under controlled conflict.

\paragraph{Accuracy and codebook adherence.}
Codebook-based measurement research shows that definitions and other coding guidance affect predictions and downstream estimates \citep{halterman2026codebook,halterman2026protest}. Recent studies more directly distinguish accuracy from responsiveness to supplied definitions: \citet{he2026} examine richer guidance and reassigned codebook labels, while \citet{casanova2026} study how models' prior concept knowledge interacts with additional instructions and examples; related perturbation studies also find inconsistent adherence to external definitions \citep{mohammadi2025llms}. We extend this work by asking whether fine-tuning increases adherence to reassigned definitions and how that adherence changes when we independently vary definition and demonstration mappings while holding retrieved texts fixed.

\paragraph{Retrieved demonstrations and label mappings.}
Retrieving examples similar to the text being classified can improve in-context classification, including political-text and many-category tasks \citep{liu2022,milios2023,liu2024}. What models learn from those examples is less clear. Randomizing demonstration labels sometimes has limited effects, but models can also learn novel mappings from demonstrations and lose accuracy when input--label pairs are corrupted \citep{min2022,wei2023,milios2023}. Moreover, nearby labeled examples can support classification without an LLM \citep{cover1967,milios2023}. We therefore compare retrieval strategies and benchmark the LLM against direct \(k\)-nearest-neighbor classification using the same neighborhood. Whereas prior retrieval studies generally evaluate performance under the original label mapping, we manipulate their labels and cross demonstration and definition mappings to examine responses to conflicting guidance.

\section{Methods}

\subsection{Conceptual Framework}
\label{sec:framework}

A codebook combines output labels with category entries. An LLM may draw on the entries' substantive content, prior associations with labels, and labeled demonstrations; standard performance cannot separate these sources when they encode the same mapping. Throughout, \emph{definition} denotes a category entry's substantive content, including any clarifications and codebook-provided examples; \emph{codebook information} denotes the complete label-and-entry information supplied in a prompt. \emph{Codebook adherence} means following the active definition--label assignment, with \emph{definition adherence} denoting the corresponding audit outcome. We call a labeled training item an \emph{example} before retrieval and a \emph{demonstration} once it is placed in the prompt.

We separate the three sources by reassigning definitions, comparing retrieval architectures with direct nearest-neighbor prediction, and independently manipulating definition and demonstration mappings. Applying the audit before and after fine-tuning tests whether higher standard performance entails greater definition adherence.

\subsection{Counterfactual Codebook Audit}
\label{sec:m-audit}

Let \(\{(\ell_j,d_j)\}_{j=1}^{J}\) denote the label--definition units included in an audit. We construct a counterfactual codebook by assigning each definition \(d_j\) to a different label:
\[
\mathcal{D}_{\pi}
=
\{(\ell_{\pi(j)},d_j):j=1,\ldots,J\},
\]
where \(\pi\) is a derangement, so \(\pi(j)\neq j\) for every category. For a text from human-labeled category \(j\), the original mapping supports \(\ell_j\), whereas the reassigned definition supports \(\ell_{\pi(j)}\). The entry remains unchanged; only its attached label is replaced.

We classify a prediction as \emph{Definition} if it equals \(\ell_{\pi(j)}\), \emph{Original} if it equals \(\ell_j\), \emph{Other} if it is another valid output label, and \emph{Invalid} otherwise. These outcomes identify which mapping a prediction matches, not the information source that produced it. The intervention is diagnostic rather than a proposed alternative taxonomy.

We repeat the audit after task-specific fine-tuning; paired changes in definition and original-label agreement distinguish responsiveness to the supplied mapping from stronger reproduction of the original one.

\subsection{Example Retrieval and Mechanism Analysis}
\label{sec:m-exrag}

\emph{Global Example-RAG} retrieves the \(k\) nearest training examples and presents them as a shared demonstration block, producing a query-specific neighborhood that may concentrate on few categories. \emph{Class-wise Example-RAG} retrieves within each category and attaches the demonstrations to its codebook entry, guaranteeing coverage but potentially including weakly relevant texts. Their comparison tests query-specific against category-balanced retrieval.

We then compare Global Example-RAG with direct \(k\)-nearest-neighbor classification using the same retrieved examples and examine neighborhood label composition, testing how much gain is already encoded in the local input--label structure.

\subsection{Interventions on Demonstration Labels}
\label{sec:m-rekey}

Demonstrations may help through their texts or their input--label associations. We distinguish these mechanisms by relabeling fixed retrieved texts.

\paragraph{Pairwise label swaps.}
We swap a frequent label pair while holding the query, retrieved texts, order, and codebook fixed. Movement toward the substituted labels shows influence beyond the texts' semantic content. Comparing prompts with and without codebook information tests whether the full codebook prompt reduces this influence; because these are complete prompt conditions, we interpret the contrast as a prompt-level rather than a pure definition effect.

\paragraph{Crossed definition and demonstration mappings.}
We independently vary the mappings encoded by definitions and demonstration labels. Category label names retain their original forms in all conditions. Table~\ref{tab:counterfactual-design} summarizes the mappings used in this work. 

\begin{table}[t]
\centering
\small
\setlength{\tabcolsep}{4pt}
\begin{tabular}{lll}
\toprule
\textbf{Condition} & \textbf{Definitions} & \textbf{Demo labels} \\
\midrule
Original               & Original   & Original \\
Reassigned definitions & Reassigned & -- \\
Reassigned demos       & --         & Reassigned \\
Definition conflict    & Reassigned & Original \\
Demo conflict          & Original   & Reassigned \\
Aligned                & Reassigned & Reassigned \\
\bottomrule
\end{tabular}
\caption{Conditions crossing the mappings encoded by definitions and demonstrations. A dash indicates that the corresponding source is not supplied.}
\label{tab:counterfactual-design}
\end{table}

The single-source conditions assess reassigned definitions and demonstrations separately against the original label names. \emph{Definition conflict} combines reassigned definitions with original demonstration labels; \emph{Demo conflict} combines original definitions with reassigned demonstration labels; and \emph{Aligned} makes both sources support the reassigned mapping.

Comparing \emph{Definition conflict} with \emph{Aligned} isolates the mapping encoded by demonstration labels: the definitions and retrieved texts remain fixed, while the demonstration labels change from the original to the reassigned mapping. Increased reassigned-mapping agreement under \emph{Aligned} would show that consistently relabeled demonstrations can help implement the supplied definition mapping.

Comparing \emph{Demo conflict} with \emph{Reassigned demos} instead tests whether adding original codebook information restores original-label agreement when demonstrations support the reassigned mapping. This is a definition-specific comparison only when the output-label menu and task instructions are otherwise identical; if they differ, it estimates the effect of adding the complete codebook prompt. Comparing \emph{Demo conflict} with \emph{Aligned} more directly isolates the mapping encoded by definitions because both conditions contain the same reassigned demonstrations and differ only in the definition--label mapping.

\section{Experimental Setup}
\label{sec:setup}

\subsection{Datasets}
\label{sec:data}

We use the human-labeled training, development, and test splits of three political-text datasets prepared by \citet{halterman2026codebook}. \textbf{CCC} contains U.S. news reports of protest events \citep{crowdcounting2024}; \textbf{BFRS}, reports of political violence in Pakistan categorized partly by actors, targets, and forms of violence \citep{bueno2015bfrs}; and \textbf{Manifesto}, party-platform sentences classified by policy \citep{lehmann2017manifesto}. We retain all eight CCC and 12 BFRS categories and, after collapsing duplicate output labels, 49 Manifesto categories (Table~\ref{tab:data}).

Training examples support retrieval and fine-tuning; development data select retrieval depth, and test data provide final evaluation.

\subsection{Models and Prompt Construction}
\label{sec:m-models}

Mistral is the primary model. The fine-tuning audit uses Mistral-7B-Instruct-v0.2 and Llama-3.1-8B-Instruct; retrieval and crossed-mapping analyses also use Llama-3.3-70B to vary family and scale. Appendix~\ref{app:implementation} gives model versions, decoding, and chat templates; Tables~\ref{tab:models} and~\ref{tab:crossmodel} report additional open-weight models.

Prompt conditions provide labels only, codebook information only, demonstrations only, labels with demonstrations, or codebook information with the same demonstrations. CCC and BFRS use their full codebooks. Manifesto label-only prompts list all 49 retained labels, while codebook prompts generally present the 30 retained entries most similar to the query. Controlled comparisons that vary codebook information while holding the available labels fixed use all 49 Manifesto entries. Dataset descriptions, Manifesto context, and output instructions remain fixed (Appendix~\ref{app:prompts}).

\subsection{Audit and Fine-Tuning Implementation}
\label{sec:m-audit-setup}

The zero-shot audit presents the retained codebook without demonstrations. We evaluate ten independent rearrangements on 1,000 documents per dataset; supporting 200-document analyses use one rearrangement (Appendix~\ref{app:audit}).

For each fine-tuning model, we train one LoRA adapter per dataset on the human-labeled training set \citep{hu2022lora}, yielding six adapters. Base and fine-tuned models use the same documents and rearrangements. All adapters share training settings and are distinct from standard-performance baselines (Appendix~\ref{app:audit}).

\subsection{Retrieval and Intervention Implementation}
\label{sec:m-retrieval-setup}

After removing training texts duplicated in the evaluation split, we rank training examples by cosine similarity using \texttt{all-mpnet-base-v2}; labels are attached only after retrieval. On development data, Mistral selects \(k=8\), 10, and 20 for Manifesto, CCC, and BFRS, fixed thereafter. Direct \(k\)NN predicts the modal retrieved label, breaking ties by summed similarity.

Class-wise Example-RAG retrieves two demonstrations per represented category (98, 10, and 24 for Manifesto, CCC, and BFRS). Matched-budget Global conditions separate retrieval strategy from count. Manifesto retrieval is restricted to its 49 retained labels. Additional retrieval, leakage, prompt-placement, and constrained-decoding controls appear in Appendix~\ref{app:exrag}.

The pairwise intervention swaps \textsc{welfare\_positive}/\textsc{environmentalism\_positive} (Manifesto), \textsc{protest}/\textsc{rally} (CCC), and \textsc{assassination}/\textsc{terrorism} (BFRS). Retrieval precedes relabeling, and neighbors are reused across crossed-mapping conditions. Paired comparisons hold fixed the query, texts, order, count, category order, label menu, instructions, decoding, and parser except for the manipulated component. For the supporting full-reassignment analysis, we test five rearrangements with Mistral and one with Llama-3.3-70B; additional controls appear in Appendix~\ref{app:paired}.

\subsection{Evaluation and Uncertainty}
\label{sec:m-eval}

Standard performance is weighted \(F_1\) against human labels; Appendix Table~\ref{tab:macro-new} reports macro-\(F_1\). Counterfactual \(F_1\) uses the supplied mapping. The audit also reports Definition, Original, Other, and Invalid outcomes; in crossed conditions, \emph{reassigned-mapping agreement} replaces definition adherence when demonstrations may support \(\ell_{\pi(j)}\).

For pairwise swaps, we report weighted-\(F_1\) loss and the rate at which baseline predictions in the swapped pair change to the paired label.

All direct comparisons use the same evaluation documents. Retrieval analyses use 500 document-level bootstrap resamples for individual scores and 2,000 paired resamples for differences, with common draws across conditions.

The fine-tuning audit uses 5,000 hierarchical paired-bootstrap replicates, resampling rearrangements and then documents within them with common draws across checkpoints. Counterfactual \(F_1\) is calculated over valid outputs; invalid outputs are reported separately.

\section{Results}
\label{sec:results}

\subsection{Original-Label Agreement Exceeds Definition Adherence}
\label{sec:r-audit}

\begin{table}[t]
\centering
\small
\setlength{\tabcolsep}{3pt}

\begin{tabular}{lrrrr}
\toprule
\textbf{Dataset} & \textbf{Definition} & \textbf{Original} & \textbf{Other} & \textbf{Invalid} \\
\midrule
CCC        & 10.75 (4.0) & 55.80 (6.4) & 32.35 (6.9)& 1.10 (0.5) \\
BFRS       & 12.87 (6.5) & 20.93 (4.0) & 65.34 (6.1) & 0.86 (0.9) \\
Manifesto & 3.24 (1.9)  & 30.25 (2.5) & 57.94 (2.3) & 8.57 (1.3)  \\
\bottomrule
\end{tabular}%
\caption{Zero-shot audit outcomes for Mistral-7B, in percent, across ten definition--label rearrangements (mean with standard deviation in parentheses; 1,000 documents per dataset).}
\label{tab:permutation-robustness}
\end{table}

To test whether models follow supplied definitions when they conflict with the original mapping, we reassign definitions among output labels. Under reassignment, original-label agreement exceeds definition adherence on all three datasets: 55.80\% versus 10.75\% on CCC, 20.93\% versus 12.87\% on BFRS, and 30.25\% versus 3.24\% on Manifesto (Table~\ref{tab:permutation-robustness}). Predictions therefore more often preserve the original mapping; definition adherence varies across rearrangements, particularly on BFRS, but remains low on average.

CCC has the highest original-label agreement, potentially reflecting its small label space and informative names such as \emph{rally} and \emph{march}. BFRS has the lowest original-label agreement and most valid predictions matching neither mapping. Manifesto has the lowest definition adherence and highest invalid rate, consistent with selection among 49 labels.

\subsection{Fine-Tuning Reinforces the Original Mapping, Not Definition Adherence}
\label{sec:r-ft}

\begin{table}[t]
\centering
\small
\setlength{\tabcolsep}{3pt}
\begin{tabular}{@{}lrr@{}}
\toprule
& \textbf{$\Delta$ Definition} & \textbf{$\Delta$ Original} \\
\midrule
\multicolumn{3}{@{}l}{\emph{Mistral-7B}} \\
CCC        & $-$3.92 [$-$8.13, $-$0.25] & $+$10.81 [$+$6.90, $+$15.41] \\
BFRS       & $-$0.18 [$-$3.20, $+$2.84] & $+$29.44 [$+$24.13, $+$34.46] \\
Manifesto  & $-$0.12 [$-$0.99, $+$0.73] & $+$7.78 [$+$5.89, $+$9.60] \\
\midrule
  \multicolumn{3}{@{}l}{\emph{Llama-3.1-8B}} \\
CCC        & $-$2.97 [$-$4.80, $-$1.21] & $+$9.93 [$+$7.99, $+$11.88] \\
BFRS       & $+$1.15 [$-$3.96, $+$5.73] & $+$18.96 [$+$14.39, $+$23.14] \\
Manifesto  & $+$1.96 [$+$0.73, $+$3.19] & $+$4.63 [$+$2.54, $+$6.70] \\
\bottomrule
\end{tabular}%
\caption{Paired percentage-point changes from base to fine-tuned models across ten definition--label rearrangements, with 95\% confidence intervals in brackets.}
\label{tab:ft-summary}
\end{table}

We compare base and LoRA-fine-tuned models on the same 1,000 documents under the same ten rearrangements to test whether fine-tuning increases responsiveness to supplied definitions.

Fine-tuning increases original-label agreement in all six model--dataset combinations (4.63--29.44 points) and all 60 rearrangement-level comparisons (Table~\ref{tab:ft-summary}). Definition adherence decreases on CCC for both models, changes unreliably in three combinations, and increases only for Llama on Manifesto; even there, its 1.96-point gain is smaller than the 4.63-point original-label gain.

Standard weighted \(F_1\) also improves (Appendix~\ref{app:audit}); better standard performance can therefore reflect reproduction of original annotations without greater definition adherence. The audit tests adoption of a revised label--definition mapping, not new category boundaries.

\subsection{Global Retrieval Gains Largely Reflect Neighborhood Labels}
\label{sec:r-exrag}
\label{sec:r-mech}

To assess the effect and mechanism of retrieval, we compare Global and Class-wise Example-RAG and then analyze Global retrieval in greater detail. Global retrieval is evaluated with demonstrations alone and with codebook information. All conditions retain the \emph{Original} mapping in Table~\ref{tab:counterfactual-design}.

\begin{table}[t]
\centering
\setlength{\tabcolsep}{2pt}
\small

\begin{tabular}{lccc}
\toprule
\textbf{Prompt information} & \textbf{Manifesto} & \textbf{CCC} & \textbf{BFRS} \\
 & \footnotesize dev/test & \footnotesize dev/test & \footnotesize dev/test \\
\midrule
\multicolumn{4}{l}{\emph{Mistral-7B}} \\
Label only                  & $.394/.394$ & $.647/.665$ & $.441/.449$ \\
Codebook only               & $.416/.404$ & $.639/.658$ & $.545/.548$ \\
Global: demo only           & $.563/\mathbf{.555}$ & $\mathbf{.696}/\mathbf{.710}$ & $.782/\mathbf{.778}$ \\
Global: codebook+demo       & $\mathbf{.580}/.551$ & $.691/.677$ & $\mathbf{.794}/.774$ \\
Class-wise:  & &  &  \\
codebook+demo   & $.538/.513$ & $.667/.668$ & $.738/.719$ \\
\midrule
\multicolumn{4}{l}{\emph{Llama-3.3-70B}} \\
Label only                  & $.465/.443$ & $.665/.664$ & $.591/.591$ \\
Codebook only               & $.495/.485$ & $\mathbf{.682}/.652$ & $.707/.704$ \\
Global: demo only           & $\mathbf{.615}/\mathbf{.610}$ & $.680/\mathbf{.685}$ & $.812/.805$ \\
Global: codebook+demo       & $.583/.570$ & $.641/.639$ & $\mathbf{.813}/\mathbf{.806}$ \\
Class-wise: & &  &  \\
codebook+demo   & $.543/.527$ & $.644/.644$ & $.773/.767$ \\
\bottomrule
\end{tabular}%

\caption{Development/test weighted \(F_1\) under the original mapping. Global conditions use 8, 10, and 20 demonstrations for Manifesto,
CCC, and BFRS, respectively. Class-wise retrieval uses 98, 10,
and 24 demonstrations. Matched-count Global comparisons appear
in Table~\ref{tab:perclass}. Bold marks the best condition within each model, dataset, and split.}
\label{tab:headline}
\end{table}

\paragraph{Retrieval architecture and performance.}

Retrieval generally improves weighted \(F_1\), most on Manifesto and BFRS. Adding Global demonstrations to Mistral's codebook prompt raises development/test \(F_1\) from .416/.404 to .580/.551 on Manifesto and from .545/.548 to .794/.774 on BFRS (Table~\ref{tab:headline}); CCC gains are smaller and model-dependent. Random demonstrations reduce \(F_1\) in all 12 model--dataset--split comparisons, showing that relevant retrieval, rather than demonstrations alone, drives the gains (Appendix~\ref{app:additional-controls}). Macro-\(F_1\) gains are uneven across categories, particularly on CCC (Appendix Table~\ref{tab:macro-new}).

Despite using more demonstrations, Class-wise retrieval underperforms Global demo-only in all 12 comparisons and Global codebook+demo in 10; the other differences are at most .005. At matched demonstration counts, Global retrieval also significantly outperforms Class-wise retrieval on Manifesto and BFRS for both models and splits (Appendix Table~\ref{tab:perclass}). We therefore focus on Global retrieval.

\paragraph{Predictive information in the retrieved neighborhood.}

Global demo-only outperforms codebook-only prompting in 11 of 12 paired bootstrap comparisons and exceeds label-only prompting numerically in all 12 (Appendix Figure~\ref{fig:steps}). Adding label names or codebook information after retrieval has smaller, sometimes negative effects, suggesting that the neighborhood supplies most of the improvement.

\begin{figure}[!t]
\centering
\includegraphics[width=\columnwidth]{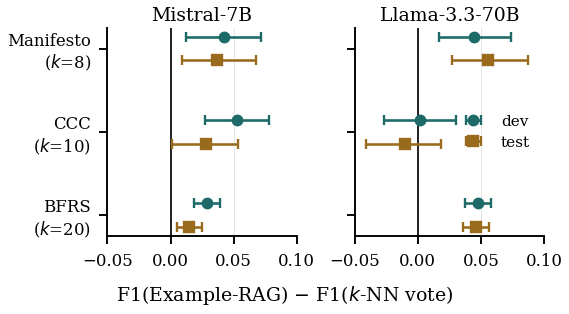}
\caption{Weighted-\(F_1\) difference between Global codebook-plus-demonstrations prompting and direct \(k\)NN using the same demonstrations. Positive values favor the LLM prompt; error bars show paired 95\% confidence intervals. The numeric results are in Table~\ref{tab:knn}}
\label{fig:knn}
\end{figure}

Direct \(k\)NN using the same retrieved examples recovers 74\%/75\% of Mistral's Global codebook+demo gain on Manifesto development/test and 88\%/93\% on BFRS, relative to codebook-only. We omit CCC because its small, prompt-dependent gain makes the ratio unstable. Global codebook+demo exceeds \(k\)NN in five of six model–dataset pairs, but only by .015--.056 weighted \(F_1\); the exception is Llama on CCC (Figure~\ref{fig:knn}). Thus, neighborhood labels explain most of the gain on Manifesto and BFRS, with a smaller contribution from the complete LLM prompt.

The modal label averages 52.5\%, 70.0\%, and 75.9\% of retrieved demonstrations on Manifesto, CCC, and BFRS, providing a strong local predictive signal. The LLM's advantage is largest when leading labels have similar similarity support; with unanimous labels, \(k\)NN matches or exceeds the LLMs (Appendix~\ref{app:retrieval-diagnostics}).

\subsection{Definitions and Demonstrations Compete Under Conflict}
\label{sec:r-conflict}
\label{sec:r-rekey}

We compare the conditions defined in Table~\ref{tab:counterfactual-design} to determine which mapping governs predictions when definitions, demonstrations, and category-label associations disagree.

\begin{table}[t]
\centering
\small
\setlength{\tabcolsep}{3pt}
\begin{tabular}{@{}lccccc@{}}
\toprule
& \multicolumn{3}{c}{\textbf{Original target}}
& \multicolumn{2}{c}{\textbf{Reassigned target}} \\
\cmidrule(lr){2-4}\cmidrule(lr){5-6}
\textbf{Dataset}
& \textbf{Orig.}
& \shortstack{\textbf{Demo}\\\textbf{conflict}}
& \shortstack{\textbf{Reassigned}\\\textbf{demos}}
& \shortstack{\textbf{Def.}\\\textbf{conflict}}
& \textbf{Aligned} \\
\midrule
\multicolumn{6}{@{}l}{\emph{Mistral-7B}} \\
Manifesto & .551 & .466 & .439 & .415 & .529 \\
CCC       & .677 & .515 & .488 & .221 & .567 \\
BFRS      & .774 & .455 & .407 & .370 & .739 \\
\midrule
\multicolumn{6}{@{}l}{\emph{Llama-70B}} \\
Manifesto & .570 & .543 & .538 & .433 & .563 \\
CCC       & .639 & .566 & .594 & .279 & .597 \\
BFRS      & .806 & .430 & .402 & .392 & .793 \\
\bottomrule
\end{tabular}
\caption{Test weighted \(F_1\) under original, conflicting, and aligned mappings. Targets follow the column group; Table~\ref{tab:counterfactual-design} defines the conditions. \emph{Reassigned demo} omits codebook information. Retrieved texts remain fixed.}
\label{tab:conflict}
\end{table}

\paragraph{Codebook information partially counters conflicting demonstrations.}

With labels reassigned away from the original mapping, \emph{Reassigned demos} yields test \(F_1\) of .402--.594, below the corresponding Global labels-demonstrations scores in all six comparisons (Table~\ref{tab:paired-full}). Adding the original codebook prompt (\emph{Demo conflict}) recovers .005--.048 in five comparisons but loses .028 for Llama on CCC (Table~\ref{tab:conflict}). \emph{Demo conflict} remains .027--.376 below \emph{Original}, so codebook information only partially counters conflicting demonstration labels.

\paragraph{Alignment supports the reassigned mapping.}
Under \emph{Definition conflict}, reassigned-target \(F_1\) is low. Relabeling the same demonstrations to match the definitions (\emph{Aligned}) increases test \(F_1\) by .114--.401, with all paired differences significant.

Alignment nevertheless does not fully recover performance under the original mapping. Across development and test sets, \emph{Aligned} remains significantly below \emph{Original} in ten of twelve comparisons and is statistically indistinguishable in the other two. Alignment substantially improves performance under the reassigned mapping, but usually does not recover performance under the original mapping.

When every definition receives a different label, Mistral's mean test \(F_1\) rises from .021--.025 under \emph{Definition conflict} to .485--.704 under \emph{Aligned}, significantly for every reassignment; Llama shows the same direction under its single rearrangement (Appendix~\ref{app:paired}).

Overall, demonstrations strongly influence the followed mapping: the original codebook only partially offsets reassigned labels, whereas alignment sharply increases adherence to the new mapping.

\section{Discussion}

\paragraph{Accuracy does not establish codebook adherence.}
Across datasets, models more often preserve the original mapping than follow reassigned definitions. Fine-tuning consistently increases original-label agreement but does
not consistently increase definition adherence. Better standard performance can therefore reflect reproduction of historical annotations rather than application of the active codebook. This inference rests on behavioral agreement with competing mappings. The audit distinguishes which mapping predictions match but does not identify their internal cause. Its counterfactual mappings serve as diagnostic tests rather than proposed taxonomies.

\paragraph{Retrieval primarily supplies local supervision.}
Global Example-RAG generally outperforms Class-wise Example-RAG, including at matched demonstration counts, favoring a focused, query-specific neighborhood over broad category coverage. This comparison does not isolate relevance from label concentration because the architectures change both. Direct \(k\)NN recovers 74--93\% of the gain on Manifesto and BFRS, showing that neighborhood labels provide much of the predictive signal. The LLM's smaller contribution is concentrated in neighborhoods with competing labels. Retrieval can therefore improve accuracy by supplying a local input--label mapping rather than improving definition use.

\paragraph{Definitions and demonstrations exert competing influence.}
Reassigned demonstration labels redirect predictions, and the original codebook only partially counters this effect. Aligning
demonstrations with reassigned definitions sharply improves performance under the new mapping, although it usually remains below performance under the original mapping. Codebook information can therefore counter conflicting demonstrations without consistently overriding them. Definitions are therefore neither ignored nor uniformly authoritative: their influence depends on whether demonstrations instantiate the same mapping.

\paragraph{Implications for LLM-assisted measurement.}
The label system, codebook, demonstration bank, retrieval procedure, and fine-tuning data jointly constitute the measurement instrument. Changing only definitions may leave the model following an earlier policy encoded elsewhere. Researchers should therefore evaluate these components jointly, supplementing standard accuracy checks with controlled tests of how predictions respond when definition and demonstration-label mappings are varied independently. Counterfactual auditing provides a stress test before model-generated annotations become substantive measurements.

\section{Conclusion}

LLM annotation should be evaluated as a joint measurement system in which written definitions, category labels, demonstrations, and training data all shape predictions. Our counterfactual audit separates definition adherence from preservation of a dataset's original mapping. Across three tasks, models rarely followed reassigned definitions; fine-tuning reinforced the original mapping, and retrieval gains largely reflected neighborhood labels. Relabeling demonstrations redirected predictions, while aligning them with definitions increased adherence. For social science measurement, predictive accuracy and responsiveness to the researcher’s coding rules are therefore distinct evaluation goals. Counterfactual auditing makes this distinction testable by placing normally aligned sources of guidance in controlled conflict, helping researchers assess what conventional accuracy measures leave unresolved.

\section*{Limitations}

We study three English-language political-text datasets and selected open-weight models; other models may balance definitions, labels, and demonstrations differently. 
Reassignment preserves category meanings, so it tests responsiveness to a supplied label--definition mapping rather than learning new category boundaries. Agreement also does not reveal internal causal mechanisms. Our claims concern observable behavior under controlled conflicts that hold queries, retrieved texts, ordering, decoding, and parsing fixed. 

Our conflict tests use predefined label swaps and reassignments while holding category content and retrieved texts fixed. They isolate responses to disagreement between mappings, but do not establish how models respond to isolated mislabeled examples or ambiguities in the wording of codebook entries.

\bibliography{combined_references}

\appendix

\makeatletter
\setlength{\@dblfptop}{0pt}
\setlength{\@dblfpsep}{12pt}
\setlength{\@dblfpbot}{0pt plus 1fil}
\makeatother

\section{Implementation Details}
\label{app:implementation}

\paragraph{Models.}
The retrieval analyses use Mistral-7B-Instruct-v0.2 and Llama-3.3-70B-Instruct. We also evaluate Llama-3.1-8B-Instruct and Gemma-7B-it under the three prompt conditions in Table~\ref{tab:models}. The fine-tuning audit uses Mistral-7B-Instruct-v0.2 and Llama-3.1-8B-Instruct. Additional zero-shot audit models are Llama-3-8B, Gemma-7B, and Llama-3.3-70B (Table~\ref{tab:crossmodel}). Because Llama-3-8B differs from Llama-3.1-8B, its zero-shot results are not a matched baseline for the adapters \citep{grattafiori2024llama3}.

\paragraph{Decoding and prompt assembly.}
For the retrieval analyses, all models use 4-bit (NF4) precision \citep{dettmers2023qlora} and greedy decoding, with responses limited to 30 new tokens. The bootstrap intervals therefore capture variation across evaluation documents rather than variation from response sampling. Prompts contain a system message and a user message formatted with the model's chat template. Llama-3.3-70B retains a separate system turn; Mistral places the system message at the beginning of the user turn (Appendix~\ref{app:prompts}). The user message presents label information, preceding context (Manifesto only), source information, retrieved examples ordered by decreasing similarity, the document, and a final instruction to return only a label.

\paragraph{Parsing.}
The parser extracts the label following \texttt{Label:}. Responses that cannot be matched to an available label are recorded as Invalid. Table~\ref{tab:constrained} reports these rates for free-text responses and compares predictive performance with generation restricted to valid labels.

\section{Audit Setup}
\label{app:audit}

\paragraph{Audit documents and label sets.}
After removing rows with missing labels, we select the first rows of each development file and use the same documents for every model, checkpoint, and rearrangement. The label distributions from first 1,000 rows for three datasets differ from the full development files by total variation distances of 0.000, 0.042, and 0.026, respectively. For the first 200 rows, used for Table~\ref{tab:crossmodel}, the corresponding distances are 0.052, 0.076, and 0.107. The main comparisons use 1,000 rows to obtain broader category coverage. 

\paragraph{Rearrangements.}
Each rearrangement assigns every codebook entry to a different label, so no entry retains its original label. Definitions, clarifications, and codebook-provided examples remain together and unchanged; only the label attached to the entry changes. We generate ten independent rearrangements and use the same ten for every model and checkpoint.

\paragraph{Fine-tuning configuration.}
The six adapters are trained only on human-labeled training rows: 4,710 for CCC, 20,978 for BFRS, and 5,755 for Manifesto. Training uses cross-entropy loss, LoRA rank 8 with alpha equal to the rank, dropout 0.05, query and value projections, a learning rate of $3\times10^{-5}$, one epoch with an effective batch size of 32, and 8-bit paged AdamW. All adapters use the same script and attention implementation. We evaluate base and fine-tuned models on the same 1,000 documents under the same ten rearrangements. Confidence intervals use 5,000 hierarchical paired-bootstrap resamples over rearrangements and documents, as described in the main text.

\paragraph{Performance under the original codebook.}
Fine-tuning improves weighted F1 under the original codebook in all six model–dataset combinations.

\paragraph{Additional zero-shot audit models.}
Table~\ref{tab:crossmodel} extends the zero-shot audit to three additional models on the 200-document sample under one rearrangement.

\begin{table*}[!t]
\centering
\begin{tabular}{llrrrr}
\toprule
Dataset & Outcome & Mistral-7B & Llama-3-8B & Gemma-7B & Llama-3.3-70B  \\
\midrule
CCC & Definition & 16.5 & 28.0 & 17.5 & 24.5  \\
    & Original   & 56.0 & 45.5 & 29.0 & 57.0  \\
    & Other      & 27.5 & 26.5 & 53.5 & 18.5 \\
    & Invalid    & 0.0  & 0.0  & 0.0  & 0.0   \\
\midrule
BFRS & Definition & 6.0  & 18.0 & 4.0  & 24.5  \\
     & Original   & 15.5 & 32.5 & 8.5  & 45.0  \\
     & Other      & 77.0 & 49.0 & 87.5 & 30.5  \\
     & Invalid    & 1.5  & 0.5  & 0.0  & 0.0   \\
\midrule
Manifesto & Definition & 3.0  & 7.0  & 2.5  & 0.0   \\
          & Original   & 28.5 & 31.5 & 18.0 & 49.5  \\
          & Other      & 60.0 & 61.0 & 36.0 & 50.5  \\
          & Invalid    & 8.5  & 0.5  & 43.5 & 0.0   \\
\bottomrule
\end{tabular}
\caption{Zero-shot audit outcomes for four models, as percentages of 200 documents under one rearrangement. Definition denotes agreement with the reassigned target; Original denotes agreement with the original human label; Other denotes another valid label; Invalid denotes an unparseable response. }
\label{tab:crossmodel}
\end{table*}

\section{Retrieval Setup and Controls}
\label{app:exrag}

\paragraph{Data accounting.}
Table~\ref{tab:data} reports usable labeled rows and the training pools available for retrieval. Manifesto uses 49 distinct output labels. We exclude two unlabeled CCC development rows and retain the supplied human annotations without reannotating the documents. Some identical training texts have conflicting labels: seven on Manifesto, forty on CCC, and five on BFRS.

\begin{table*}[!t]
\centering
\begin{tabular}{lrrr}
\toprule
 & Manifesto & CCC & BFRS \\
\midrule
Document is a & sentence & article & report \\
Median words per document & $17$ & $444$ & $27$ \\
\midrule
Codebook entries & $142$ & $8$ & $12$ \\
Labels ever used in the data & $55$ & $5$ & $12$ \\
Labels scored & $49$ & $5$ & $12$ \\
\midrule
Train rows, usable & $5{,}755$ & $4{,}710$ & $20{,}978$ \\
Dev rows, usable & $1{,}247$ & $1{,}007$ & $4{,}495$ \\
Test rows, usable & $1{,}217$ & $1{,}010$ & $4{,}496$ \\
\midrule
Example pool, dev runs & $5{,}711$ & $4{,}709$ & $20{,}909$ \\
Example pool, test runs & $5{,}717$ & $4{,}710$ & $20{,}930$ \\
\bottomrule
\end{tabular}
\caption{Dataset sizes and retrieval pools. Usable rows have human labels. Manifesto is scored on 49 retained output labels; CCC and BFRS use all observed labels. Retrieval pools contain usable training rows, restricted on Manifesto to those 49 labels, after removing texts also present in the evaluation split.}
\label{tab:data}
\end{table*}

\begin{figure*}[!t]
\centering
\includegraphics[width=\textwidth]{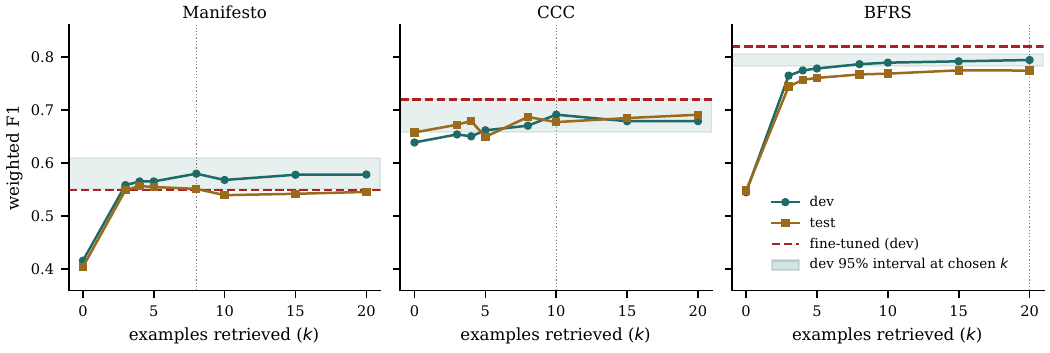}
\caption{Weighted F1 by the number of retrieved examples $k$ for Mistral-7B with codebook information; $k{=}0$ supplies no examples. Dotted vertical lines mark the development-selected values used for test evaluation. Shading shows the 95\% bootstrap interval for development F1 at the selected value. Beyond $k{=}8$, development F1 varies by at most 0.021 within each dataset.}
\label{fig:ksweep}
\end{figure*}

\paragraph{Splits, leakage filtering, and codebook coverage.}
We select $k$ on development data and hold it fixed for test evaluation. Exact-overlap filtering removes 23/1/69 training rows for Manifesto/CCC/BFRS development evaluation and 17/0/48 for test evaluation. Table~\ref{tab:selection-controls} evaluates additional similarity filtering and, for Manifesto, exclusion of examples from the query's own manifesto. The Manifesto 30-entry codebook contains the reference category for 92.9\% of development and 93.6\% of test documents. Replacing the 30-entry codebook with the full retained codebook
significantly changes Mistral's F1 only in the development condition without examples (0.377 versus 0.416). CCC and BFRS prompts include all codebook entries. Figure~\ref{fig:ksweep} reports the development comparison used to select $k$.

\begin{table*}[!t]
\centering
\setlength{\tabcolsep}{4pt}
\begin{tabular}{lcccccccc}
\toprule
Dataset & $k{=}0$ & $k{=}3$ & $k{=}4$ & $k{=}5$ & $k{=}8$ & $k{=}10$ & $k{=}15$ & $k{=}20$ \\
\midrule
Manifesto & .416/.404 & .559$^*$/.549 & .566/.557 & .566/.555 & \textbf{.580/.551} & .568/.540 & .578/.542 & .579/.546 \\
CCC & .639/.658 & .654$^*$/.672 & .650$^*$/.679 & .662$^*$/.649$^*$ & .671$^*$/.687 & \textbf{.691/.677} & .679/.685 & .679/.691 \\
BFRS & .545/.548 & .765$^*$/.744$^*$ & .775$^*$/.757$^*$ & .778$^*$/.761$^*$ & .786$^*$/.767$^*$ & .789/.769 & .792/.775 & \textbf{.794/.774} \\
\bottomrule
\end{tabular}
\caption{Values plotted in Figure~\ref{fig:ksweep}}
\label{tab:ksweep}
\end{table*}

\paragraph{All prompt conditions.}

\begin{table*}[!t]
\centering
\begin{tabular}{lccc}
\toprule
Condition & Manifesto & CCC & BFRS \\
 & dev/test & dev/test & dev/test \\
\midrule
\multicolumn{4}{l}{Mistral-7B} \\
1 Label only & $.394$/$.394$ & $.647$/$.665$ & $.441$/$.449$ \\
2 Codebook only & $.416$/$.404$ & $.639$/$.658$ & $.545$/$.548$ \\
3 Global: demo only & $.563$/$.555$ & $.696$/$.710$ & $.782$/$.778$ \\
4 Global: labels+demo & $.518$/$.519$ & $.711$/$.703$ & $.764$/$.757$ \\
5 Global: codebook+demo & $.580$/$.551$ & $.691$/$.677$ & $.794$/$.774$ \\
6 Class-wise: codebook+demo & $.538$/$.513$ & $.667$/$.668$ & $.738$/$.719$ \\
\midrule
\multicolumn{4}{l}{Llama-3.3-70B} \\
1 Label only & $.465$/$.443$ & $.665$/$.664$ & $.591$/$.591$ \\
2 Codebook only & $.495$/$.485$ & $.682$/$.652$ & $.707$/$.704$ \\
3 Global: demo only & $.615$/$.610$ & $.680$/$.685$ & $.812$/$.805$ \\
4 Global: labels+demo & $.611$/$.612$ & $.684$/$.689$ & $.813$/$.808$ \\
5 Global: codebook+demo & $.583$/$.570$ & $.641$/$.639$ & $.813$/$.806$ \\
6 Class-wise: codebook+demo & $.543$/$.527$ & $.644$/$.644$ & $.773$/$.767$ \\
\bottomrule
\end{tabular}
\caption{Development/test weighted F1 under the original mapping for all six prompt conditions. Conditions 3 to 5 use the selected $k$ (8/10/20); condition 6 supplies two examples per represented category (98/10/24). Individual 95\% confidence intervals are approximately $\pm0.03$ on Manifesto and CCC and $\pm0.012$ on BFRS. Paired bootstrap comparisons appear in Figure~\ref{fig:steps}.}
\label{tab:ladder}
\end{table*}

\begin{figure*}[!htp]
\centering
\includegraphics[width=\textwidth]{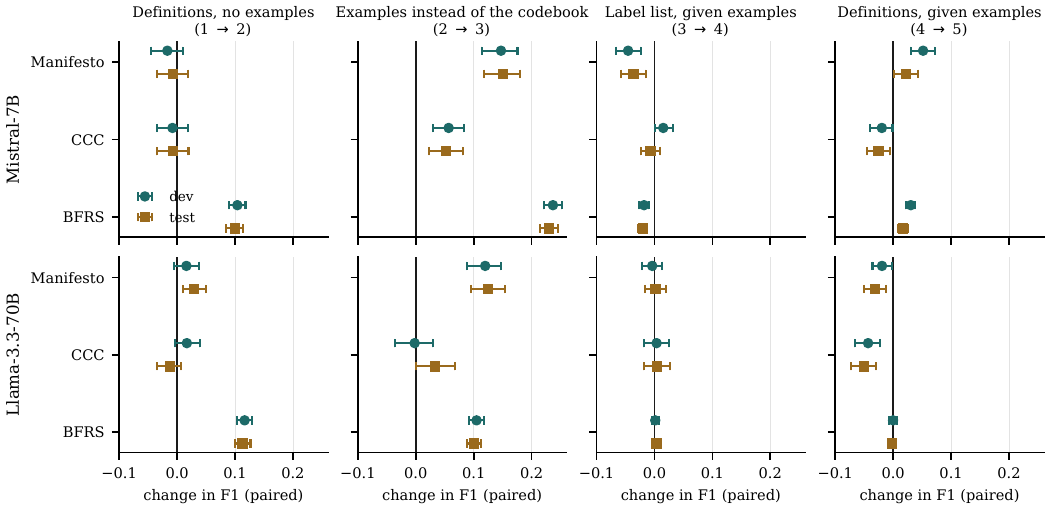}
\caption{Paired differences in weighted F1 for four prompt comparisons. Circles show development-set results and squares test-set results; positive values favor the later condition. Conditions are numbered as in Table~\ref{tab:ladder}. The Manifesto comparisons that add codebook information use the
same 49 output labels in both conditions.}
\label{fig:steps}
\end{figure*}

Table~\ref{tab:ladder} reports all six prompt conditions for both main models. The main-text headline table presents conditions 1, 2, 3, 5, and 6; the additional condition supplies label names with demonstrations. Figure~\ref{fig:steps} reports paired differences between prompt conditions. The two Manifesto comparisons that add codebook information use the
full retained label set on both sides rather than the top-30
configuration in Table~\ref{tab:ladder}. Comparing codebook-only with demo-only prompting both removes codebook information and adds demonstrations. The instruction sentence also changes to describe the supplied information, so this contrast does not isolate a definition effect.

\newpage
\begin{table*}[!htp]
\centering
\begin{tabular}{lcc}
\toprule
Comparison, dataset & Dev [95\% CI] & Test [95\% CI] \\
\midrule
\multicolumn{3}{l}{\emph{Mistral-7B}} \\
\multicolumn{3}{l}{1$\rightarrow$2: add codebook, no demos} \\
\quad Manifesto$^\dagger$ & $-.017$ [$-.045$, $+.010$] & $-.007$ [$-.034$, $+.019$] \\
\quad CCC & $-.008$ [$-.035$, $+.019$] & $-.007$ [$-.035$, $+.020$] \\
\quad BFRS & $+.104^*$ [$+.090$, $+.118$] & $+.100^*$ [$+.085$, $+.114$] \\
\multicolumn{3}{l}{2$\rightarrow$3: demos instead of codebook} \\
\quad Manifesto & $+.147^*$ [$+.114$, $+.176$] & $+.151^*$ [$+.118$, $+.181$] \\
\quad CCC & $+.057^*$ [$+.030$, $+.083$] & $+.052^*$ [$+.024$, $+.082$] \\
\quad BFRS & $+.237^*$ [$+.221$, $+.252$] & $+.230^*$ [$+.214$, $+.245$] \\
\multicolumn{3}{l}{3$\rightarrow$4: add label list to demos} \\
\quad Manifesto & $-.046^*$ [$-.066$, $-.024$] & $-.036^*$ [$-.058$, $-.015$] \\
\quad CCC & $+.015^*$ [$+.000$, $+.032$] & $-.007$ [$-.023$, $+.010$] \\
\quad BFRS & $-.018^*$ [$-.026$, $-.010$] & $-.020^*$ [$-.028$, $-.013$] \\
\multicolumn{3}{l}{4$\rightarrow$5: add codebook to labels+demos} \\
\quad Manifesto$^\dagger$ & $+.052^*$ [$+.030$, $+.072$] & $+.022^*$ [$+.002$, $+.043$] \\
\quad CCC & $-.020^*$ [$-.040$, $-.001$] & $-.026^*$ [$-.044$, $-.006$] \\
\quad BFRS & $+.030^*$ [$+.022$, $+.038$] & $+.017^*$ [$+.008$, $+.025$] \\
\midrule
\multicolumn{3}{l}{\emph{Llama-3.3-70B}} \\
\multicolumn{3}{l}{1$\rightarrow$2: add codebook, no demos} \\
\quad Manifesto$^\dagger$ & $+.016$ [$-.005$, $+.038$] & $+.029^*$ [$+.010$, $+.050$] \\
\quad CCC & $+.017$ [$-.004$, $+.039$] & $-.013$ [$-.034$, $+.007$] \\
\quad BFRS & $+.117^*$ [$+.104$, $+.130$] & $+.113^*$ [$+.100$, $+.127$] \\
\multicolumn{3}{l}{2$\rightarrow$3: demos instead of codebook} \\
\quad Manifesto & $+.120^*$ [$+.088$, $+.147$] & $+.125^*$ [$+.095$, $+.154$] \\
\quad CCC & $-.002$ [$-.036$, $+.029$] & $+.034^*$ [$+.000$, $+.068$] \\
\quad BFRS & $+.105^*$ [$+.093$, $+.117$] & $+.100^*$ [$+.088$, $+.113$] \\
\multicolumn{3}{l}{3$\rightarrow$4: add label list to demos} \\
\quad Manifesto & $-.004$ [$-.021$, $+.013$] & $+.002$ [$-.016$, $+.020$] \\
\quad CCC & $+.004$ [$-.019$, $+.026$] & $+.004$ [$-.018$, $+.027$] \\
\quad BFRS & $+.001$ [$-.005$, $+.007$] & $+.003$ [$-.003$, $+.009$] \\
\multicolumn{3}{l}{4$\rightarrow$5: add codebook to labels+demos} \\
\quad Manifesto$^\dagger$ & $-.019^*$ [$-.036$, $-.002$] & $-.031^*$ [$-.050$, $-.012$] \\
\quad CCC & $-.043^*$ [$-.066$, $-.022$] & $-.051^*$ [$-.072$, $-.029$] \\
\quad BFRS & $-.000$ [$-.007$, $+.006$] & $-.002$ [$-.009$, $+.005$] \\
\bottomrule
\end{tabular}
\caption{Values plotted in Figure~\ref{fig:steps}: paired change in weighted F1 from the earlier to the later prompt condition, numbered as in Table~\ref{tab:ladder}, with 95\% paired bootstrap intervals (2,000 resamples). $^*$ marks an interval excluding zero. $^\dagger$ marks Manifesto comparisons in which the codebook side uses all entries, so both conditions show the same labels.}
\label{tab:steps}
\end{table*}

\newpage

\paragraph{Global versus Class-wise at matched example counts.}
At matched example counts, Global retrieval significantly outperforms Class-wise retrieval on Manifesto and BFRS for both models and splits (Table~\ref{tab:perclass}). Holding the example count fixed shows that the advantage on Manifesto and BFRS is not explained by demonstration count, although relevance and label composition still vary together.

\begin{table*}[!t]
\centering
\begin{tabular}{lccc}
\toprule
 & Manifesto & CCC & BFRS \\
 & dev/test & dev/test & dev/test \\
\midrule
Examples in prompt & $98$ & $10$ & $24$ \\
\midrule
\multicolumn{4}{l}{Mistral-7B} \\
Class-wise & $.538^*$/$.513^*$ & $.667^*$/$.668$ & $.738^*$/$.719^*$ \\
Global, same count & $.567$/$.543$ & $.691$/$.677$ & $.794$/$.782$ \\
\midrule
\multicolumn{4}{l}{Llama-3.3-70B} \\
Class-wise & $.543^*$/$.527^*$ & $.644$/$.644$ & $.773^*$/$.767^*$ \\
Global, same count & $.597$/$.582$ & $.641$/$.639$ & $.819$/$.808$ \\
\bottomrule
\end{tabular}
\caption{Class-wise Example-RAG against Global Example-RAG with the same number of examples; both prompts include codebook entries. Weighted F1, development/test. $^*$ marks a Class-wise score whose paired difference from Global has a 95\% bootstrap interval excluding zero; every starred difference favors Global.}
\label{tab:perclass}
\end{table*}

\paragraph{Additional models.}
Table~\ref{tab:models} evaluates three prompt conditions for Llama-3.1-8B and Gemma-7B alongside the two main models. Demo-only prompting produces higher F1 than label-only prompting for every model and dataset. Adding codebook information to demonstrations does not significantly improve performance on Manifesto or CCC.

\begin{table*}[!t]
\centering
\begin{tabular}{lccc}
\toprule
Prompt & Manifesto & CCC & BFRS \\
 & dev/test & dev/test & dev/test \\
\midrule
\multicolumn{4}{l}{Mistral-7B} \\
Label only & $.394$/$.394$ & $.647$/$.665$ & $.441$/$.449$ \\
Global: demo only & $.563$/$.555$ & $.696$/$.710$ & $.782$/$.778$ \\
Global: codebook+demo & $.580$/$.551$ & $.691$/$.677^*$ & $.794^*$/$.774$ \\
\midrule
\multicolumn{4}{l}{Llama-3.1-8B} \\
Label only & $.402$/$.378$ & $.618$/$.582$ & $.461$/$.462$ \\
Global: demo only & $.533$/$.533$ & $.665$/$.678$ & $.770$/$.770$ \\
Global: codebook+demo & $.489^*$/$.484^*$ & $.658$/$.668$ & $.774$/$.764$ \\
\midrule
\multicolumn{4}{l}{Gemma-7B} \\
Label only & $.263$/$.263$ & $.598$/$.619$ & $.224$/$.207$ \\
Global: demo only & $.582$/$.571$ & $.608$/$.620$ & $.777$/$.771$ \\
Global: codebook+demo & $.570$/$.550^*$ & $.589$/$.585^*$ & $.766^*$/$.764$ \\
\midrule
\multicolumn{4}{l}{Llama-3.3-70B} \\
Label only & $.465$/$.443$ & $.665$/$.664$ & $.591$/$.591$ \\
Global: demo only & $.615$/$.610$ & $.680$/$.685$ & $.812$/$.805$ \\
Global: codebook+demo & $.583^*$/$.570^*$ & $.641^*$/$.639^*$ & $.813$/$.806$ \\
\bottomrule
\end{tabular}
\caption{Three prompts on four models, weighted F1, development/test, $k$ = 8/10/20. $^*$ marks a codebook+demo score that differs from the same model's demo-only score (95\% paired bootstrap interval excludes zero). Models are Mistral-7B-Instruct-v0.2, Llama-3.1-8B-Instruct, Gemma-7B-it, and Llama-3.3-70B-Instruct.}
\label{tab:models}
\end{table*}

\paragraph{Constrained decoding.}
Table~\ref{tab:constrained} compares free-text generation with decoding restricted to valid labels. Across 48 recomputed prompt comparisons, no difference that is significant under free generation reverses direction, although six change statistical significance. For the larger model, adding codebook information to demo-only prompts still significantly lowers performance on the Manifesto and CCC test sets, but the development differences are no longer significant. Constrained decoding can change responses that would otherwise produce valid labels, so its effect extends beyond preventing parsing failures.

\begin{table*}[!t]
\centering
\begin{tabular}{lccc}
\toprule
 & Manifesto & CCC & BFRS \\
 & dev/test & dev/test & dev/test \\
\midrule
\multicolumn{4}{l}{Free-text answers that fail to parse (\%), Mistral} \\
1 Label only & $5.1$/$6.6$ & $0.9$/$1.0$ & $4.1$/$3.9$ \\
2 Codebook only & $6.0$/$5.6$ & $0.8$/$0.9$ & $5.1$/$4.7$ \\
3 Demo only & $15.9$/$16.4$ & $2.8$/$1.3$ & $3.5$/$3.8$ \\
4 Labels+demo & $3.9$/$4.7$ & $0.5$/$0.0$ & $1.4$/$1.3$ \\
5 Codebook+demo & $3.8$/$2.9$ & $0.3$/$0.2$ & $0.6$/$0.6$ \\
\midrule
\multicolumn{4}{l}{Constrained minus free, weighted F1, Mistral} \\
1 Label only & $-.010^*$/$-.016^*$ & $-.003$/$-.003$ & $+.004^*$/$+.002$ \\
2 Codebook only & $+.003$/$.000$ & $+.006^*$/$-.004^*$ & $-.006^*$/$-.006^*$ \\
3 Demo only & $-.005$/$-.002$ & $+.009^*$/$.000$ & $+.002$/$+.003$ \\
4 Labels+demo & $-.006$/$-.008^*$ & $+.003$/$-.004^*$ & $.000$/$.000$ \\
5 Codebook+demo & $-.001$/$-.005^*$ & $+.001$/$-.001$ & $.000$/$.000$ \\
\midrule
\multicolumn{4}{l}{Constrained minus free, weighted F1, Llama-3.3-70B} \\
1 Label only & $.000$/$+.006$ & $-.002$/$-.002$ & $.000$/$.000$ \\
2 Codebook only & $+.003$/$+.009$ & $-.019^*$/$-.028^*$ & $.000$/$.000$ \\
3 Demo only & $-.002$/$-.001$ & $+.022^*$/$+.026^*$ & $.000$/$.000$ \\
4 Labels+demo & $+.002$/$-.007$ & $+.018^*$/$+.026^*$ & $.000$/$-.001$ \\
5 Codebook+demo & $+.014^*$/$+.010$ & $+.040^*$/$+.029^*$ & $.000$/$.000$ \\
\bottomrule
\end{tabular}
\caption{Constrained decoding against free text on the same documents. Top: share of free-text answers from which no valid label can be read; constrained decoding makes this zero. Middle and bottom: weighted F1 with decoding restricted to valid labels minus weighted F1 with free text. $^*$ marks a difference whose 95\% paired bootstrap interval excludes zero. Conditions are numbered as in Table~\ref{tab:ladder}.}
\label{tab:constrained}
\end{table*}

\paragraph{Macro-F1.}
Table~\ref{tab:macro-new} reports macro-F1, giving each category equal weight. On CCC, codebook-only prompting exceeds codebook+demo prompting for both models. For Mistral, this reverses the ordering under weighted F1. The low retrieval coverage of rare categories in Appendix~\ref{app:retrieval-diagnostics} helps explain why the two metrics give different assessments.

\begin{table*}[!t]
\centering
\begin{tabular}{lccc}
\toprule
 & Manifesto & CCC & BFRS \\
 & dev/test & dev/test & dev/test \\
\midrule
\multicolumn{4}{l}{Mistral-7B} \\
Codebook only & $.318$/$.283$ & $.382$/$.437$ & $.453$/$.438$ \\
Demo only & $.439$/$.432$ & $.357$/$.472$ & $.724$/$.706$ \\
Codebook+demo & $.457$/$.396$ & $.343$/$.430$ & $.732$/$.696$ \\
\midrule
\multicolumn{4}{l}{Llama-3.3-70B} \\
Codebook only & $.387$/$.346$ & $.462$/$.480$ & $.667$/$.650$ \\
Demo only & $.526$/$.498$ & $.343$/$.451$ & $.768$/$.756$ \\
Codebook+demo & $.496$/$.442$ & $.309$/$.411$ & $.770$/$.759$ \\
\midrule
Direct $k$NN vote & $.420$/$.390$ & $.306$/$.381$ & $.679$/$.671$ \\
\bottomrule
\end{tabular}
\caption{Macro-F1, development/test, giving equal weight to each category. Demo only and codebook+demo are the Global conditions. The direct $k$NN vote uses the same retrieved examples as the Global conditions.}
\label{tab:macro-new}
\end{table*}

\subsection{Example-Selection Controls}
\label{app:additional-controls}

Random selection uses the same example count as nearest-neighbor retrieval; balanced random selection also distributes examples across categories. Similarity filtering excludes examples with query cosine similarity at or above 0.85 (changes at a 0.90 threshold are at most 0.005). Source filtering, available only for Manifesto, excludes all training examples from the query's manifesto. Table~\ref{tab:selection-controls} reports these controls. Random selection significantly lowers F1 in all twelve model--dataset--split comparisons. Balanced random selection also lowers F1 throughout Manifesto and BFRS, but CCC differences are not consistently significant or negative. Changes under the leakage filters are at most 0.008, except when excluding examples from the same manifesto. The larger decline under source filtering indicates that part of the retrieval gain depends on examples from the query's source.

\begin{table*}[!t]
\centering
\begin{tabular}{lrrrrr}
\toprule
 & & & Balanced & Remove & Remove \\
Dataset, model & Nearest & Random & random & sim.\ $\geq .85$ & same source \\
\midrule
Manifesto, Mistral & .580/.551 & .432*/.396* & .414*/.382* & .575*/.549* & .518*/.483* \\
CCC, Mistral & .691/.677 & .602*/.615* & .626*/.659 & .685/.680 & --- \\
BFRS, Mistral & .794/.774 & .642*/.635* & .611*/.623* & .789*/.769* & --- \\
Manifesto, 70B & .583/.570 & .485*/.490* & .482*/.481* & .581/.568 & .543*/.524* \\
CCC, 70B & .641/.639 & .606*/.610* & .638/.646 & .637/.634 & --- \\
BFRS, 70B & .813/.806 & .735*/.732* & .704*/.703* & .806*/.804 & --- \\
\bottomrule
\end{tabular}
\caption{Development/test weighted F1 for example-selection controls with codebook information. Stars mark paired differences from nearest-neighbor retrieval whose bootstrap confidence intervals exclude zero.}
\label{tab:selection-controls}
\end{table*}

\section{Retrieval Coverage and the Contribution of the Language Model}
\label{app:retrieval-diagnostics}

At the selected $k$, development neighborhoods contain an average of 4.09 distinct labels on Manifesto (of 49), 3.02 on BFRS (of 12), and 2.46 on CCC (of 5). The most frequent label accounts for 52.5\%, 75.9\%, and 70.0\% of retrieved examples, respectively; test shares differ by at most 0.1 points. Retrieved labels thus concentrate on a small set of categories, although the model can output other labels. The reference label appears in the neighborhood for 84.1/83.5\% of Manifesto, 94.1/95.8\% of CCC, and 97.1/97.0\% of BFRS development/test documents. Coverage is lower for rare categories: it is 66\% for Manifesto development documents in the rarest third of categories and 12\% for CCC DEMONSTRATION documents. This uneven coverage helps explain differences between weighted F1 and macro-F1.

\begin{table*}[!t]
\centering
\begin{tabular}{lllccc}
\toprule
Model & Dataset & Split & LLM & $k$NN vote & Difference [95\% CI] \\
\midrule
Mistral-7B & Manifesto & dev & .580 & .538 & $+.042^*$ [$+.012$, $+.071$] \\
 &  & test & .551 & .515 & $+.037^*$ [$+.009$, $+.068$] \\
 & CCC & dev & .691 & .639 & $+.053^*$ [$+.028$, $+.078$] \\
 &  & test & .677 & .649 & $+.029^*$ [$+.001$, $+.054$] \\
 & BFRS & dev & .794 & .765 & $+.029^*$ [$+.018$, $+.040$] \\
 &  & test & .774 & .760 & $+.015^*$ [$+.005$, $+.025$] \\
\midrule
Llama-3.3-70B & Manifesto & dev & .583 & .538 & $+.045^*$ [$+.017$, $+.074$] \\
 &  & test & .570 & .515 & $+.056^*$ [$+.027$, $+.087$] \\
 & CCC & dev & .641 & .639 & $+.002$ [$-.027$, $+.030$] \\
 &  & test & .639 & .649 & $-.010$ [$-.041$, $+.018$] \\
 & BFRS & dev & .813 & .765 & $+.048^*$ [$+.037$, $+.058$] \\
 &  & test & .806 & .760 & $+.046^*$ [$+.036$, $+.056$] \\
\bottomrule
\end{tabular}
\caption{Values plotted in Figure~\ref{fig:knn}.}
\label{tab:knn}
\end{table*}

Table~\ref{tab:vote-subsets} compares the Global codebook+demo prompt with the direct $k$NN vote in development subsets. The vote matches or exceeds both models when retrieved labels are unanimous. Both models have higher accuracy when neighborhoods contain at least four labels or the two leading labels have similar summed similarity. These descriptive comparisons locate the LLM's additional predictive value in neighborhoods where labels compete; they do not establish a causal mechanism.

\begin{table*}[!t]
\centering
\begin{tabular}{lrrr}
\toprule
 & Unanimous & $\geq 4$ labels & Closest support \\
Dataset, model & LLM / vote & LLM / vote & LLM / vote \\
\midrule
Manifesto, Mistral & 83.1 / 93.8 & 48.5 / 41.9 & 47.6 / 28.9 \\
Manifesto, 70B & 89.2 / 93.8 & 50.7 / 41.9 & 49.2 / 28.9 \\
CCC, Mistral & 79.4 / 79.4 & 71.2 / 61.0 & 66.7 / 52.4 \\
CCC, 70B & 79.4 / 79.4 & 69.5 / 61.0 & 59.5 / 52.4 \\
BFRS, Mistral & 96.9 / 97.7 & 66.8 / 61.9 & 58.4 / 49.3 \\
BFRS, 70B & 96.9 / 97.7 & 71.4 / 61.9 & 63.4 / 49.3 \\
\bottomrule
\end{tabular}
\caption{Development-set accuracy (\%) of the Global codebook+demo prompt and of the direct $k$NN vote within retrieved-example subsets. Closest support is the quarter of documents with the smallest difference in summed similarity between the two leading labels. These are descriptive subsets without subgroup significance tests.}
\label{tab:vote-subsets}
\end{table*}

\section{Paired Changes to Codebook and Example Labels}
\label{app:paired}

The paired experiment uses the full retrieval development and test sets, fixed retrieved texts and order, and the selected example counts (8/10/20). Manifesto uses the 30-entry codebook from the corresponding original prompt; CCC and BFRS use their full codebooks. Table~\ref{tab:counterfactual-design} defines the conditions. Reassigned demos supplies reassigned demonstrations with a label list and no definitions; for Manifesto, this list includes all 49 label names, whereas codebook prompts show 30 entries. The contrast between Reassigned demos and Demo conflict therefore changes the complete prompt, including the available label information. Table~\ref{tab:swap-loss} separately reports switch rates for a demo-only prompt without a label list and for Demo conflict. A neutral-name control replaces familiar category names with identifiers under Definition conflict.

Weighted F1 uses the original human label as the target for Original, Demo conflict, and Reassigned demos, and the reassigned target for Definition conflict, Reassigned definitions, Aligned, and the neutral-name control. Documents whose human label belongs to the swapped pair account for about 18\%, 87\%, and 51\% of the Manifesto, CCC, and BFRS development sets. These shares differ from the switch-rate denominator in Table~\ref{tab:swap-loss}, which selects documents by the model's prediction before the swap.

\begin{table*}[!t]
\centering
\setlength{\tabcolsep}{3.5pt}
\begin{tabular}{lcccccccccccc}
\toprule
 & \multicolumn{4}{c}{Manifesto} & \multicolumn{4}{c}{CCC} & \multicolumn{4}{c}{BFRS} \\
\cmidrule(lr){2-5}\cmidrule(lr){6-9}\cmidrule(lr){10-13}
 & \multicolumn{2}{c}{Mistral} & \multicolumn{2}{c}{70B} & \multicolumn{2}{c}{Mistral} & \multicolumn{2}{c}{70B} & \multicolumn{2}{c}{Mistral} & \multicolumn{2}{c}{70B} \\
Condition & dev & test & dev & test & dev & test & dev & test & dev & test & dev & test \\
\midrule
\multicolumn{13}{l}{Original codebook} \\
Original & .580 & .551 & .583 & .570 & .691 & .677 & .641 & .639 & .794 & .774 & .813 & .806 \\
Demo conflict & .491 & .466 & .565 & .543 & .511 & .515 & .557 & .566 & .476 & .455 & .430 & .430 \\
\midrule
\multicolumn{13}{l}{Reassigned codebook} \\
Reassigned definitions & .349 & .325 & .443 & .420 & .334 & .349 & .498 & .470 & .255 & .245 & .553 & .562 \\
Definition conflict & .452 & .415 & .472 & .433 & .224 & .221 & .292 & .279 & .380 & .370 & .399 & .392 \\
Aligned & .539 & .529 & .584 & .563 & .570 & .567 & .605 & .597 & .752 & .739 & .801 & .793 \\
Def.\ conflict, neutral names & .454 & .416 & .575 & .538 & .482 & .480 & .485 & .454 & .404 & .400 & .422 & .418 \\
\midrule
\multicolumn{13}{l}{Label list, no codebook} \\
Original demos & .518 & .519 & .611 & .612 & .711 & .703 & .684 & .689 & .764 & .757 & .813 & .808 \\
Reassigned demos & .457 & .439 & .549 & .538 & .455 & .488 & .589 & .594 & .414 & .407 & .403 & .402 \\
\bottomrule
\end{tabular}
\caption{Development/test weighted F1 under the two-label reassignment. Conditions follow Table~\ref{tab:counterfactual-design}; neutral names replaces familiar category names under Definition conflict. Original and Demo conflict are scored against original human labels; reassigned-codebook conditions are scored against the reassigned target. Label-list conditions are scored against original human labels, and swapping their demonstration labels significantly lowers performance except for the 70B Manifesto comparisons. Aligned significantly exceeds Definition conflict in all twelve comparisons and falls significantly below Original except for the two 70B Manifesto comparisons. The table extends the five conditions in Table~\ref{tab:conflict} with development results and additional controls.}
\label{tab:paired-full}
\end{table*}

\begin{table*}[!t]
\centering
\begin{tabular}{lccc}
\toprule
 & Manifesto & CCC & BFRS \\
Prompt & dev/test & dev/test & dev/test \\
\midrule
Mistral, swapped demos only & 44/45 & 66/64 & 70/70 \\
Mistral, codebook + swapped demos (Demo conflict) & 23/26 & 44/44 & 64/65 \\
Llama-3.3-70B, swapped demos only & 67/71 & 66/69 & 91/89 \\
Llama-3.3-70B, codebook + swapped demos (Demo conflict) & 9/9 & 23/20 & 80/79 \\
\bottomrule
\end{tabular}
\caption{Development/test switch rates after the pairwise label swap (\%). The denominator contains documents predicted as either swapped label before the intervention; the numerator counts predictions that move to the paired label. Swapped demos only supplies demonstrations without a label list or codebook. Demo conflict supplies the original codebook with swapped demonstrations. Weighted-F1 losses for Manifesto, CCC, and BFRS are .125/.127, .316/.326, and .364/.374 for Mistral and .133/.149, .272/.314, and .413/.407 for the 70B under swapped demos only. Corresponding losses with the codebook are .089/.086, .180/.162, .318/.319 and .018/.027, .083/.072, .384/.376; every loss is statistically significant.}
\label{tab:swap-loss}
\end{table*}

\begin{table*}[!t]
\centering
\begin{tabular}{lcccc}
\toprule
 & \multicolumn{2}{c}{Original codebook} & \multicolumn{2}{c}{Reassigned codebook} \\
\cmidrule(lr){2-3}\cmidrule(lr){4-5}
 & & Demo & Definition & \\
Dataset, model & Original & conflict & conflict & Aligned \\
\midrule
Manifesto, Mistral & .580/.551 & .267/.263 & .022/.021 & .512/.485 \\
CCC, Mistral & .691/.677 & .452/.451 & .028/.023 & .501/.497 \\
BFRS, Mistral & .794/.774 & .119/.119 & .022/.025 & .705/.704 \\
Manifesto, 70B & .583/.570 & .461/.447 & .185/.190 & .574/.536 \\
CCC, 70B & .641/.639 & .644/.626 & .040/.027 & .601/.584 \\
BFRS, 70B & .813/.806 & .199/.200 & .021/.023 & .748/.745 \\
\bottomrule
\end{tabular}
\caption{Development/test weighted F1 when every codebook entry receives a different label. Original repeats the unchanged Global codebook+demo scores. Mistral scores average five reassignments; the 70B uses one.}
\label{tab:all-reassigned}
\end{table*}

\clearpage
\section{Prompts}
\label{app:prompts}

Listing~\ref{lst:prompt} reproduces a retrieval prompt with the lengthy codebook entries summarized. Listing~\ref{lst:conditions} shows how the six prompt conditions differ. The listing text preserves the wording used in the experiment, including grammatical errors.

\begin{listing*}[tp]
\lstset{basicstyle=\scriptsize\ttfamily, numbers=none, xleftmargin=0pt,
        literate={•}{{\textbullet}}1}
\begin{lstlisting}
[SYSTEM]
Instructions: You're an expert political scientist classifying statements from political parties' manifestos into categories. Carefully read the definitions below, read the sentence from the manifesto, and write the single Label that best match the story. Do not write anything else except the Label in all caps. It's very important that you use only the provided Labels.
Task Type: single-label multi-class

[USER]
Categories:
<< 30 category blocks, not reproduced here. Each block gives Category, Aggregation1,
   Label, Definition and, where the codebook has them, Clarification and Negative
   Clarification. They are the 30 entries most similar to the document, most
   similar first; MARKET_REGULATION, the correct label, is the 6th. >>

To provide some context for your coding, here's the previous passage: "This will be a growing economy where everyone benefits - not just the richest 1%. Investment and planning will provide opportunities for everybody to have a well-paid job and a good work-life balance.".

The following is an excerpt from a political party in United Kingdom..

Here are some labeled examples from this codebook:

Example 1:
Text: Plaid Cymru wants a successful economy for Wales, where everybody can maximise their potential.
Label: ECONOMIC_GROWTH_POSITIVE

Example 2:
Text: Plaid Cymru wants to get more people in Wales into work, with better pay and conditions.
Label: LABOUR_GROUPS_POSITIVE

Example 3:
Text: Plaid Cymru believes that everybody should pay a fair share of tax so that we in Wales can all share excellent, world-class public services.
Label: EQUALITY_POSITIVE

Example 4:
Text: So that wealth is shared out more fairly within the UK, and to ensure that Wales receives its fair share of public investment, Plaid Cymru will introduce an Economic Fairness Bill to rebalance the wealth of the UK.
Label: MARKET_REGULATION

Example 5:
Text: Plaid Cymru believes in our people, and we want to give them the best chances in life.
Label: EQUALITY_POSITIVE

Example 6:
Text: Plaid Cymru wants a Wales where everybody can get involved.
Label: CIVIC_MINDEDNESS_POSITIVE

Example 7:
Text: Plaid Cymru wants all of our children and young people to reach their full potential.
Label: CIVIC_MINDEDNESS_POSITIVE

Example 8:
Text: Plaid Cymru will stand up for the needs of pensioners.
Label: WELFARE_POSITIVE

Document: As the economy improves, Plaid Cymru believes that everyone should benefit.

Write the name of the Label that fits best, with no other text. It's very important that you use only the provided labels. For example, "Label: FOREIGN_SPECIAL_POSITIVE", "Label: CONTROLLED_ECONOMY", etc.
\end{lstlisting}
\caption{Global codebook+demo prompt (condition 5) for the first Manifesto development sentence, with $k{=}8$ examples and 30 category entries. The experimental wording is preserved except that the 30 category blocks are summarized between \texttt{<<} and \texttt{>>}, and long lines are wrapped. \texttt{[SYSTEM]} and \texttt{[USER]} identify messages formatted with each model's chat template. The eight examples appear in decreasing similarity order, selected from 5,711 training sentences. They contain six distinct labels; the reference label, \texttt{MARKET\_REGULATION}, appears once, at rank four. Mistral instead predicted \texttt{ECONOMIC\_GROWTH\_POSITIVE}, the label of the nearest example.}
\label{lst:prompt}
\end{listing*}

\begin{listing*}[tp]
\lstset{basicstyle=\scriptsize\ttfamily, numbers=none, xleftmargin=0pt,
        breaklines=false, literate={•}{{\textbullet}}1}
\begin{lstlisting}
All six conditions send the same query. Conditions 3, 4 and 5 also send the same eight
examples. <CATS> = the 30 category blocks, <EX> = the eight examples, <context> = the
previous-passage line, <source> = the "The following is an excerpt ..." line, and
<reminder> = the closing "Write the name of the Label ..." line, all exactly as in Listing 1.
Order in every prompt: label information, context, source, examples, query, reminder.

(1) LABEL ONLY                                   (2) CODEBOOK ONLY
[SYSTEM] ... Carefully read the valid labels     [SYSTEM] ... Carefully read the definitions
         below, ...                                       below, ...
[USER]                                           [USER]
Valid labels:                                    Categories:
WELFARE_POSITIVE, ENVIRONMENTALISM_POSITIVE,     Category: free market economy
EQUALITY_POSITIVE,                               Aggregation1: Economy
TECHNOLOGY_AND_INFRASTRUCTURE,                   Label: FREE_MARKET_ECONOMY
EDUCATION_POSITIVE, LAW_AND_ORDER_POSITIVE,      Definition: Favourable mentions of the free
LAW_AND_ORDER_POSITIVE, DEMOCRACY, ...             market and free market capitalism as an
   << full retained label list >>           economic model.
<context> <source>                               Clarification: May include favourable
Document: <query>                                  references to: • Laissez-faire economy; ...
<reminder>                                       <CATS: the other 29 blocks>
                                                 <context> <source>
                                                 Document: <query>
                                                 <reminder>

(3) GLOBAL: DEMO ONLY                            (4) GLOBAL: LABELS + DEMO
[SYSTEM] ... Carefully read the labeled          [SYSTEM] ... Carefully read the valid labels
         examples below, ...                              below, ...
[USER]                                           [USER]
   << No label list and no definitions.          Valid labels:
      Label names appear only on the             WELFARE_POSITIVE, ENVIRONMENTALISM_POSITIVE,
      examples, and as the two format            ...  << full retained label list, as in (1) >>
      examples in <reminder>. >>                 <context> <source>
<context> <source>                               Here are some labeled examples from this
Here are some labeled examples from this         codebook:
codebook:                                        <EX: the same 8 examples as (3)>
<EX: 8 examples, by overall similarity>          Document: <query>
Document: <query>                                <reminder>
<reminder>

(5) GLOBAL: CODEBOOK + DEMO (= Listing 1)        (6) CLASS-WISE: CODEBOOK + DEMO
[SYSTEM] ... Carefully read the definitions      [SYSTEM] ... Carefully read the definitions
         below, ...                                       below, ...
[USER]                                           [USER]
Categories:                                      Categories:
<CATS: 30 blocks, as in (2)>                     <CATS: 30 blocks, as in (2)>
<context> <source>                               <context> <source>
Here are some labeled examples from this         Here are some labeled examples from this
codebook:                                        codebook:
<EX: the same 8 examples as (3)>                 <2 most similar examples for each of the 49
Document: <query>                                 labels in the pool: 98 in total, most
<reminder>                                        similar first, however dissimilar>
                                                 Document: <query>
                                                 <reminder>
\end{lstlisting}
\caption{All six prompt conditions for the same document; blocks shared with Listing~\ref{lst:prompt} are named rather than repeated. Between conditions only the label information and the one instruction sentence that names it change, and conditions 3, 4 and 5 share the same eight examples. Condition 3 never lists the valid labels: the only label names it shows are those on its examples and the two format examples in the closing reminder. The label list in conditions 1 and 4 is not de-duplicated, so the two labels with two codebook entries each appear twice.}
\label{lst:conditions}
\end{listing*}

\end{document}